\documentclass{llncs}

\usepackage{times}
\usepackage{graphicx}
\usepackage{latexsym}
\usepackage{url}

\def\ar{\leftarrow}

\newcommand{\as}{``}

\newcommand{\be}{\begin{em}}
\newcommand{\ee}{\end{em}}
\newcommand{\bb}{\begin{bf}}
\newcommand{\eb}{\end{bf}}
\newcommand{\I}[1]{\relax\ifmmode\mbox{\it#1}\else{\it#1}\fi}

\newcommand{\tbl}{\hspace*{12mm}}

\newcommand{\no}{not\,}

\newcommand{\rif}{~\ref}
\newcommand{\IF}{\mbox{\,:-\,}}

\newcommand{\emptySeq}{\relax\ifmmode \varepsilon\else$\varepsilon$\fi}   

\newcommand{\enables}{\mbox{\bf\ enables\ }}
\newcommand{\tto}{\mbox{\bf\ to\ }}
\newcommand{\on}{\mbox{\bf\ on\ }}
\newcommand{\tthen}{\mbox{\bf\ then\ }}
\newcommand{\iif}{\mbox{\bf\ if\ }}
\newcommand{\determinedby}{\mbox{\bf\ determinedby\ }}

\begin{document}
\begin{sloppypar}

\title{Committment-Based Data-Aware\\Multi-Agent-Contexts Systems}

\author{Stefania Costantini \institute{University of L'Aquila,
Italy, email: stefania.costantini@univaq.it}}

\maketitle


\begin{abstract}
Communication and interaction among agents have been the subject of extensive investigation since many years. Commitment-based communication, where communicating agents are seen as a debtor agent who is committed to a creditor agent to bring about something (possibly under some conditions) is now very well-established. The approach of
DACMAS (Data-Aware Commitment-based MAS) lifts commitment-related
approaches proposed in the literature from a propositional
to a first-order setting via the adoption the DRL-Lite Description Logic. Notably, DACMASs provide, beyond commitments, simple forms of inter-agent event-based communication. Yet, the aspect is missing of making a MAS able to acquire knowledge from contexts which are not agents and which are external to the MAS. This topic is coped with in Managed MCSs (Managed Multi-Context Systems), where however exchanges are among knowledge bases and not agents. In this paper, we propose the new approach of DACmMCMASs (Data-Aware Commitment-based managed Multi- Context MAS), so as to obtain a commitment-based first-order agent
system which is able to interact with heterogeneous external information sources. We show that DACmMCMASs retain the nice formal properties of the original approaches.
\end{abstract}

\section{Introduction}
\label{introduction}

Communication and interaction among agents in Multi-Agents Systems (MAS) have been the subject of extensive
investigation since many years (cf. \cite{ArtikisP08,ACL-Manifesto2013} for useful reviews and discussion). Commitments (cf. \cite{Singh91,Singh07,Singh00,Singh12} and the references therein) have been proposed as a general paradigm for agent interaction, where a commitment $C_{x,y\,}(\mathit{ant,csq})$, relates a \emph{debtor agent} $x$ to a \emph{creditor agent} $y$ where $x$ commits to bring about $\mathit{csq}$ whenever $\mathit{ant,csq}$ holds. Commitments can be created, discharged (i.e., fulfilled), canceled, released, delegated ad assigned, where this lifecycle is managed by a so-called \as commitment machine'' \cite{YolumS01}. In agent societies, commitments can be divided into base-level commitments and meta-commitments, which represent the \as norms'' of the society \cite{Singh99}. An implicit meta-commitment is called (and considered to be) a \as convention'', or a \as custom''. Commitment-based communications appears particularly suitable in Artificial Social Systems and Organizational Multi-Agent Systems where two or more business partners interact, where each partner is represented as an autonomous, reactive and proactive agent. Such systems are in fact often based upon (business) protocols \cite{Singh07} that regulate interactions among agents by specifying the massages that agents are able to exchange, and constraints on these messages (involving message order, sequences, contents, etc.). Literature about commitment-based protocols static and run-time verification is so wide as to make it impossible to mention all relevant work here. About run-time verification we mention \cite{REC2011,REC2012,ChesaniMMT13}, about static (a priori) verification we mention (among many) \cite{BaldoniBCDPS09,MarengoBBCPS11,BaldoniBMP13,Singh13}. In \cite{Singh00}, a semantics is proposed for the most common FIPA primitives \cite{FIPA2003} in terms of commitments. The objective of such a semantics is that of abstracting away from agents' \as mental states'', which are by definition private and non-observable, while in the author's perspective communication is a social phenomenon consisting of exchanges that are made public. However, despite the open criticism to the \as mentalistic'' approach, in this work semantics of FIPA primitives still refer to BDI \cite{rao,RG91} concepts of Belief, Desire and Intention stating, e.g., that in an \as Inform'' primitive whatever is communicated is also believed. 
Alternative formal proposals that
treat communicative acts in terms of commitments can be found in in \cite{ForCol2009}
and in the references therein.

In \cite{Singh12,SinghMASC} there is a strong claim toward an approach where a Multi-Agent System (MAS) is to be considered as
a distributed system including heterogeneous components which operate autonomously and need to interoperate. No reference to \as Intelligence'' or knowledge representation
and reasoning needs to be made, in the authors' view, concerning communication: a strict software-engineering orientation is embraced, where agents
are to any practical extents pieces of software managing interaction protocols. In \cite{Singh13}, the quest for a decoupling of agent design and protocol design is advocated, to be obtained by means of refactoring design techniques, so as to allow designers to deploy a new protocol for a MAS with no need of substantial modifications to the agents'code.

Though the commitment-based communication paradigm is very well-established and has proved effective in many practical applications, some
more or less explicit discussion about the pervasiveness of such a paradigm has emerged (see e.g. \cite{ArtikisP08,ACL-Manifesto2013}), i.e., whether or not are commitments really able to capture all forms of inter-agent communication, and whether or not intelligence and knowledge representation and reasoning abilities of agents, as well as the application context where they operate, can really be always and fully ignored or, however, considered to be a system parameter not to be taken into account as far as communication is concerned.

In interesting recent work \cite{MontaliCG14}, Calvanese De Giacomo and Montali study how to lift to lift the commitment-related
approaches proposed in the literature from a propositional
to a first-order setting. This with the aim of establishing 
how data maintained by the agents impact on the dynamics of MAS, 
and on the evolution of commitments. They are able to specify and verify (under some conditions) dynamic properties by means of a first-order variant of $\mu$-calculus \cite{Stirling01,CalvaneseGMP13}. The approach,
called DACMAS (Data-Aware Commitment-based MAS) is based upon the DRL-Lite Description Logic \cite{DL2003handbook}.
In this approach, a set of agents is augmented with an \emph{institutional} agent which owns a \as global TBox, that

\begin{quote}
\emph{represents the key concepts, relations and constraints characterizing
the domain in which the agents operate, so as to provide a common ground for the agent
interaction.}
\end{quote}

The institutional agent keeps track of all agents participating in the system and of the evolution of their commitments,
and is able to observe and record every interaction. Agents consist of a local ABox which is consistent with the global
TBox (though mutual consistency of the ABoxes is not required) and of a set of reactive and proactive rules, where proactive
rules model communications, and reactive rules model internal updates. Interestingly enough, in fact, agents communicate
by interchanging \emph{events}, which may then lead to commitment formation. For instance, a potential buyer
send an event to a seller in order to register as customer. When (via an internal update) the seller does so, it sends
to the buyer events representing offers. Only later, in case, e.g., the customer decides to buy, commitments are
(conditionally) created by the institutional agent (which implements the commitment machine) about, e.g., payment and delivery.
Notice that, since the institutional agent explicitly and implicitly participates in every interaction,
and also implements the commitment machine, if all communications were to be based upon commitment one would need
a meta-institutional agent, and so on in a potentially infinite regression.
This even in a setting where participating agents are strongly coupled by the unique common ontology.

Though the DACMAS approach is interesting, also for its potential for affordable verification, an important aspect is still
missing, at least in the hypothesis of a really heterogeneous and autonomous set of agents. In fact, decisions
about how to react to an event can be taken in DACMASs only
based upon querying the local or the institutional ABox.
In real-world applications, decisions such as, e.g., whether to accept a customer or whether to issue an order to a certain seller,
whether to enroll a candidate student, whether to concede a loan and at which conditions, etc. are taken after consulting
a number of information sources that can be not only internal, but also external to the agent system, and after following
a decision procedure that goes beyond simple reaction. This irrespective to concepts such as \as intelligence'' an
\as mental states'' that the agent communication community (at least as represented by \cite{ACL-Manifesto2013}) chooses to ignore:
for instance a University, in order to select among candidate
students, requires some documentation and collects additional information from various sources.
Then, it compares the candidates and makes decisions according to its internal criteria, and to decision-making algorithms 
relying upon knowledge representation and reasoning of some kind. Therefore, communication can be social, but the extent and
purposes and (social) effects of communication may still widely depend upon the agent's internal processes and upon its the interaction with other 
sources. This kind of interaction has little to do with sociality and commitments, like for instance when the University checks the candidates' secondary
school ranking in a local government's file.

In the Artificial Intelligence and Knowledge Representation field, The Multi-Context Systems (MCS) approach has been proposed 
to model such an information exchange \cite{BrewkaE07,BrewkaEF11}. In this proposal, there is a number of (information) contexts that are essentially kinds
of heterogeneous reasoning entities, which interact via so-called \emph{bridge rules}: such rules allow an entity to augment
its knowledge base by inferring new consequence from what can (and/or cannot) be concluded in the other contexts.
The MCS have evolved by the simplest form to Managed MCS (mMCS) \cite{BrewkaEFW11}, where conclusions resulting from bridge rules are
locally re-elaborated so as to avoid inconsistency and to be able to suitably incorporate the new knowledge.

In this paper, we propose to combine the approach of \cite{MontaliCG14} with mMCS into DACmMCMASs (DACmMCMASs), so as to obtain a commitment-based agent
system which is however able to interact with heterogeneous external information sources. This by augmenting the set of participating agents by a set of contexts, and by equipping agents with special communicative rules that are a variant of bridge rules. 

The paper is organized as follows: in Sections\rif{background} we provide the necessary background notions about mMCSs and DACMASs. In Section\rif{dacmmcmas} we present and illustrate, also by means of examples, the new approach of DACmMCMASs. We also discuss how nice properties of both mMCSs and DACMASs extend to the new formalization. Finally, in Section\rif{conclusions} we conclude.

\section{Background on mMCS and DACMAS}
\label{background}

\subsection{mMCS}
\label{mMCS}

(Managed) Multi-Context systems \cite{BrewkaEF11,BrewkaEFW11} aim at making it possible
to build systems that need to access multiple possibly heterogeneous data sources,
despite the variety of formalisms and languages these data sources can be based upon.
In particular, it is realistically assumed that data sources, called \as contexts'' cannot be standardized in any way, so
the objective is that of modeling the necessary information flow among contexts.
The device for doing so is constituted by \as bridge rules'', which are similar to datalog rules
with negation (cf., e.g., \cite{lpneg:survey} for a survey about datalog and the references therein for more information) 
but allow for inter-context communication, as in each element their \as body'of these rules the context from which information
is to be obtained is explicitly indicated.

In this section we will briefly revise MCSs. We will adapt terminology and definitions to our setting.
Reporting from \cite{BrewkaEF11}, a logic $L$ is a triple
$(KB_L; Cn_L; ACC_L)$, where $KB_L$ is the set of admissible
knowledge bases of $L$, which are sets of $KB$-elements (\as formulas'');
$Cn_L$ is the set of acceptable sets of consequences (in \cite{BrewkaEF11} they are
called \as belief sets'', but we prefer to abstract away from \as mentalistic'' concepts), whose elements
are data items or "facts"; 
$ACC_L : KB_L \rightarrow 2^{Cn_L}$ is a function which defines
the semantics of $L$ by assigning each knowledge-base an \as acceptable''
set of consequences.
A multi-context system (MCS) \(M = (C_1,\ldots,C_n)\) is a heterogeneous
collection of contexts $C_i = (L_i; kb_i; br_i)$ where $L_i$ is a logic,
$kb_i \in KB_{L_i}$ is a knowledge base and $br_i$ is a set of bridge
rules. Each such rule is of the following form, where the left-hand side $s$ is called the \emph{head}, also denoted as $hd(\rho)$,
the right-hand side is called the \emph{body}, also denoted as $\mathit{body}(\rho)$,
and the comma stand for conjunction.
\[s \ar (c_1: p_1), \ldots, (c_j : p_j),\no (c_{j+1} : p_{j+1}), \ldots, \no (c_m : p_m).\]

For each bridge rule included in a context $C_i$, it is required that $kb_i \cup s$  belongs to $KB_{Li}$ and, 
for every $k \leq m$, $c_k$ is a context included in $M$,
and each $p_k$ belongs to some set in $KB_{L_k}$.

The meaning is that $s$ is added to the consequences of $kb_i$ 
whenever each $p_r$, $r \leq j$, belongs to the consequences of context 
$c_r$, while instead each $p_s$, $j < s \leq m$, does not belong to the consequences of context $c_s$.
If $M = (C_1,\ldots,C_n)$ is an MCS, a data state
(\as belief state'' in the terminology of \cite{BrewkaEF11}) is a tuple $S = (S_1,\ldots, S_n)$ such that each $S_i$ is
an element of $Cn_i$. Desirable data states are those where each $S_i$ is acceptable according to $ACC_i$.
A bridge rule $\rho$ is applicable in a data
state iff for all $1 \leq i \leq j: p_i \in S_i$ and for
all $j + 1 \leq k \leq m: p_k \not\in S_k$.
Let $\mathit{app}(S)$ be the set of bridge rules which are applicable in a data state $S$.

We will now introduce managed MCS (mMCS) though in a simplified form with
respect to \cite{BrewkaEF11}: in fact, they substitute a logic
with a \as logic suite'', where one can select the desired semantics for the
given knowledge base. We define mMCS over logics, as the extension over logic suites is not needed in our setting
(thus, our formulation is in between those of \cite{BrewkaEF11} and \cite{BrewkaEFW11}).
While in standard MCSs the head $s$ of a bridge rule is simply added to the \as destination'' context's data state $kb$,
in managed MCS $kb$ is subjected to an elaboration w.r.t. $s$ according to a specific operator $op$ and to its intended semantics:
rather than simple addition, $op$ can determine, e.g. deletion of other formulas upon addition of $s$, or any
kind of elaboration and revision of $kb$ w.r.t. $s$. Formula $s$ itself can be elaborated by $op$, for instance with the aim
of making it compatible with $kb$'s format.

For a logic $L$, \(F_L = \{s \in kb\, |\, kb \in KB_L\}\) is the
set of formulas occurring in its knowledge bases.
A \emph{management base} is a set of operation names
(briefly, operations) $OP$, defining elaborations that can be 
performed on formulas, e.g., addition of, revision with,
etc. For a logic $L$ and a management base
$OP$, the set of
operational statements that can be built from $OP$ and $F_L$ is \(F^{OP}_L = \{o(s)\, |\, o \in OP, s \in F_L\}\).
The semantics of such statements is given by a management
function, which maps a set of operational
statements and a knowledge base into a modified knowledge
base. In particular, a management function over a logic $L$
and a management base $OP$ is a function 
\[\mathit{mng}: 2^{F^{OP}_L} \times KB^{L} \rightarrow  2^{KB_L} \setminus \emptyset\]
Bridge rules for context $C_i$ are as before,
except that the head now belongs to $F^{OP}_L$, and is then of the form $o(s)$. 

Semantics of (simplified) mMCS are in terms of \emph{equilibria}.
A data state $S = (S_1,\ldots, S_n)$ is an
equilibrium for an MCS \(M = (C_1,\ldots,C_n)\) iff, for $1 \leq i \leq n$, there exists $kb_i' =  mng_i(\mathit{app}(S),kbi)$ such that
$S_i \in ACC_i(kb_i')$.
Thus, an equilibrium is a global data state composed of acceptable data sets, 
one for each context, considering however inter-context communication determined by bridge rules
and the elaboration resulting from the operational statements and the management function. 

Equilibria may not exist, or may
contain inconsistent data sets (local inconsistency, w.r.t. \emph{local consistency}).
A management function is called
\emph{local consistency (lc-) preserving} iff, for every given 
management base, $kb'$ is consistent. It can be proved that
a mMCS where all management functions are lc-preserving is locally consistent.
\emph{Global consistency} would require the $S_i$'s to be consistent with each other, but this property is not required in this context.

Intuitively, a data state is an equilibrium whenever the application of a bridge rule
according to the destination context's strategy for incorporating new knowledge produces a result which is compatible
with the context's semantics for its data sets. E.g., in a data set consisting simply of positive
and negative facts simple addition of the negation of a pre-existing fact cannot result in an equilibrium
as it determines an inconsistency. Equilibrium 
can be guaranteed via an lc-preserving operator $op$ which, e.g., gives priority to newly acquired data by removing the pre-existing conflicting item.

Notice that bridge rules are intended to be applied whenever they are applicable, so inter-context communication automatically occurs via the predefined set of bridge rules, though mediated via the management function. In our proposal, as seen below bridge rules will be introduced in agents, which will be able to apply them proactively.

\subsection{DACMAS}
\label{dacmas}

We assume as known the basic concepts about Description Logic
and ontologies \cite{DL2003handbook} and in particular DLR-Lite, though no detail is needed here beyond being aware of the following aspects. (1) A TBox is a finite set of 
assertions specifying: concepts and relations; inclusion and disjunction among concepts/relations; key assertions
for relations. (2) An ABox is a finite set of assertions concerning concept and relation membership. In essence, a TBox describes
the structure of the data/knowledge, and the ABox specifies the actual data/knowledge instance. (3) In DLR-Lite, data can be queried via UCQs
(Union of Conjunctive Queries) and ECQs (Existential Conjunctive Queries): the latter are FOL (First-Order Logic) queries involving negation, conjunction and the existential quantifier, whose atoms are UCQs.

Formally, a DACMAS (Data-Aware Commitment-based Multi-Agent System) is (from \cite{MontaliCG14})
a tuple $\langle \cal{T}, \cal{E}, \cal{X}, \cal{I}, \cal{C}, \cal{B}\rangle$ where: (i) $\cal{X}$ is a finite set of agent specifications; (ii) \cal{T} is a global DLR-Lite TBox, which is common to all agents participating in the system; (iii) $\cal{I}$ is a specification for the \as institutional'' agent; (iv) $\cal{E}$ is a set of
predicates denoting events (where the predicate name is the event type, and the arity
determines the content/payload of the event); 
(v) $\cal{C}$ is a contractual
specification; (vi) and $\cal{B}$ is a Commitment Box (CBox). The global TBox lists the names of all participating agents
in connection to their specifications. Each agent has a local ABox, consistent with the global TBox, where
however the ABoxes of the various agents are not required to be mutually consistent. The institutional agent is a special agent who
is aware of every message exchanged in the system, and can query all ABoxes. In addition, it is responsible of the management
of commitments, whose concrete instances are maintained in the Commitment Box $\cal{B}$, and it does so based on the \emph{Commitment Rules} in $\cal{C}$, which define the commitment machine. An execution semantics for DACMASs is provided in \cite{MontaliCG14}, in terms of a transition system constructed by means of a suitable algorithm. 

Apart from the local ABox, each agent's specification include a (possibly empty set of): \emph{communicative rules}, which proactively determine events to be sent to other agents; \emph{update rules}, which are internal reactive rules that update the local ABox upon sending/receiving an event to/from another agent.
the other participants.

A communicative rule has the form
\[Q(r,\hat{x}) \enables EV(\hat{x}) \tto r\]
where: $Q$ is an $ECQ$ query, or precisely an $ECQ_{l}$ with \emph{location argument} $@\mathit{Ag}$ to specify the agent to which the query is directed (if omitted, the the agent queries its own ABox); $\hat{x}$ is a set of tuples representing the results of the query; $EV(\hat{x})$ is an event supported by the system,
i.e., predicate $EV$ belongs to $\cal{E}$;  $r$ is a variable, denoting an agent's name. Whenever the rule is proactively applied, 
if the query evaluates to true (i.e., if the query succeeds) then $EV(\hat{x})$ and $r$ are instantiated via one 
among the answers returned by the query, according to the agent's own choice. For instance, an agent can find the name $r$ of the provider of a service (if several names are returned, only one is chosen) and sends to this provider a subscription request (instantiated with the necessary information $\hat{x}$) in order to be able to access the service.

Update rules are ECA-like rules\footnote{As it is well-known, 'ECA' rules stands for 'Event-Condition-Action' rules, and specify reaction to events.} of the following form, where $\alpha$ is an action, 
the other elements are as before, and each rule is
to be applied whenever an event is either sent or received, as specified in the rule itself:
\[\on EV(\hat{x}) \tto r \iif Q(r,\hat{x}) \tthen \alpha(r,\hat{x}) \mbox{\bf\ (on-send/on-receive)}\]
Update rules may imply the insertion in the agent's ABox of new data items not previously present in the
system, taken from a countably infinite domain $\Delta$.
For instance, after subscription to a service an agent can receive offers and issue orders, the latter case determining the creation of a commitment (managed by the institutional agent).

An agent specification is a tuple $\langle \mathit{sn},\Pi \rangle$, where $\mathit{sn}$ is
the agent specification name, and $\Pi$ is the set of communicative and update
rules characterizing the agent.

\section{DACmMCMAS: Data-Aware Commitment-based managed Multi-Context MASDACmMCMASs}
\label{dacmmcmas}

The definition of a DACmMCMAS ( Data-Aware Commitment-based managed Multi-Context MASDACmMCMASs), extend that of DACMASs as the set of participating agents is augmented with a set of contexts, which are to be understood as data sources which can be consulted by agents.
The global TBox will now include a set of context names. However no specification of contexts is provided, since contexts are supposed to be external to the system. Contexts are supposed to be available for queries only. Each context's name is however related, in the Tbox, to a \emph{context role}, specifying the function that a context assumes in the system. E.g., a context's name $\mathit{studoff}$ may correspond to context role $\mathit{student\_office}$, and context name $\mathit{poldept}$ to $\mathit{police\_department}$. More information about contexts can be held in each agent's ABox, i.e., each agent may have its own private information about context roles. Bridge rules, similar to those of MCSs,
can be now defined also in agents. 
Each agent is therefore equipped with local management functions,
which can perform any elaboration on acquired data. The objective is to keep
the agent's ABox consistent, by means either of general simple techniques
(e.g., rejecting incoming inconsistent knowledge) or via more involved belief revision techniques.

In the following, let a logic, a management base and management functions be as specified in Section\rif{mMCS}. 
Formally, we have (where $\cal{T}, \cal{E}, \cal{X}, \cal{I}, \cal{C}$ and $\cal{B}$ are the same as for DACMASs,
their meaning is reported only for making the overall definition self-contained):

\begin{definition}
A DACmMCMAS (Data-Aware Commitment-based managed Multi- Context MAS) is
a tuple $\langle \cal{T}, \cal{E}, \cal{X}, \cal{Y}, \cal{I}, \cal{C}, \cal{B} \rangle$ where: (i) \cal{T} is a global DLR-Lite TBox, which is common to all agents participating in the system; (ii) $\cal{E}$ is a set of
predicates denoting events (where the predicate name is the event type, and the arity
determines the content/payload of the event); (iii) $\cal{X}$ is a finite set of agent specifications;
(iv) $\cal{Y}$ is a finite set of context names;
(vv) $\cal{I}$ is a specification for the institutional agent; (vi) $\cal{C}$ is a contractual
specification; (vii) and $\cal{B}$ is a Commitment Box (CBox). 
\end{definition}

The global TBox lists, as in DACMASs, the names of all participating agents
in connection to their specifications. The global TBox however also lists the description
of all available external contexts, for which (differently from agents)
no specification is available within the system. Each context has a name and, as mentioned, each context has a \emph{role}, indicating to agents
the kind of information that can be obtained from such context. For simplicity, we assume that roles are specified as constants:
in future evolution of this work, more expressive descriptions may be adopted.
We assume the each agent's local ABox may include additional context descriptions, concerning contexts which are locally known to that specific agent.
We also assume that context names include all the information needed 
for actually posing queries (e.g., context names might coincide with their URIs).
Context names might also be linked to the information about the related query language;
however, again for the sake of simplicity though without loss of generality
we assume that all contexts accept datalog queries.
In particular, we will consider datalog queries 
of the following form.

\begin{definition}
\label{dq}
An agent-to-context datalog query is defined as follows:
\[Q \mbox{\IF} A_1,\ldots,A_n, \no B_1, \no B_m \mbox{\ with\ } n+m > 0\]
where the left-hand-side $Q$ can stand in place of the right-hand-side.
The comma stand for conjunction, and each the $A_i$s is either an atom or a binary expression 
involving connectives such as equality, inequality, comparison, applied to variables
occurring in atoms and to constants. Each atom has a (possibly empty) tuple of arguments
and can be either ground, i.e., all arguments are constants, 
or non-ground, i.e, arguments include both constants and variables, to be instantiated to constants
in the query results. All variables which occur either in $Q$ or in the $B_i$s also occur in the $A_i$s. 
\end{definition}

Intuitively, the conjunction of the $A_i$s selects a set of tuples and the $B_i$s rule some of them out.
$Q$ is essentially a placeholder for the whole query, but also projects over the wished-for elements of the resulting tuples.

Each context may include bridge rules, of the form specified in section\rif{mMCS},
where however the body refers to contexts only, i.e., contexts cannot query agents.
The novelty of our approach is that also agents may be equipped with bridge rules,
for extracting data from contexts (not agents, as data exchange among agents occurs via explicit communication
as defined in DACMASs).
As in DACMASs, we assume that the new data items possibly added 
to the agent's ABox belong to the same countably infinite domain $\Delta$.
Moreover, as seen below bridge rules in agents are not automatically applied as in mMCSs,
rather they are proctively activated by agents upon need.

\begin{definition}
A bridge rule occurring in an agent's specification has the following form.

\[A(\hat{x}) \determinedby E_1,\ldots,E_k, \no G_{k+1}, \ldots, \no G_r \]

$A(\hat{x})$, called the \emph{conclusion} of the rule, is an atom over tuple of arguments $\hat{x}$. 
The right-hand-side is called the \emph{body} of the rule, and is 
a conjunction of queries on external contexts.
Precisely, each of the $E_i$s and each of the $G_i$s (where $k > 0$ and $r \geq 0$) can be either of the form $DQ_i(\hat{x_i}):c_i$
or of the form $DQ_i(\hat{x_i}):q_i$
where: $DQ_i$ is a datalog query (defined according to Definition\rif{dq})  over tuple of arguments $\hat{x_i}$;
$c_i$ is a context listed in the local ABox with is role, and thus locally known to the agent; $q_i = \mbox{Role@inst}(\mathit{role_i})$ is a context name obtained by means of a standard query $\mbox{Role@inst}$ to the institutional agent $\mbox{inst}$ (notation '@' is borrowed from standard DACMASs), performed by providing the context role $\mathit{role_i}$. We assume that all variables occurring in $A(\hat{x})$ and in each of the $G_i$s also occur in the $E_i$s. The comma stands for conjunction. Assuming (without loss of generality) that all the $\hat{x_i}$s have the same arity, when the rule is \emph{triggered} (see Definition\rif{trigger} below) then the $E_i$s may produce a set of tuples, some of which are discarded by the $G_i$s. Finally, $A(\hat{x})$ is obtained as a suitable projection.
Within an agent, different bridge rules have distinct conclusions. The management operations and function are defined separately (see Definition\rif{beca} below). 
\end{definition}

E.g., {\mbox{Role@inst(\emph{student\_office})} would return the name of the context corresponding to the student office.
There is, as mentioned, an important difference between bridge rules in contexts and bridge rules in agents.
Each bridge rules in a context is meant to be automatically applied whenever the present data state entails the rule body.
The new knowledge corresponding to the rule head is added (via the management function) to the context's knowledge base.
Instead, bridge rules in agents are meant to be proactvely activated by the agent itself.
To this aim, we introduce suitable variants of DACMAS's communicative and update rules.

\begin{definition}
\label{trigger}
A \emph{trigger rule} has the form
\[Q(\hat{x}) \enables A(\hat{y})\]
where: $Q$ is an $ECQ_{l}$ query, and $\hat{x}$ a set of tuples representing the results of the query; $A(\hat{y})$ is 
the conclusion of exactly one of the agent's bridge rules. If the query evaluates to true, then $A(\hat{y})$ is (partially) instantiated via one 
among the answers returned by the query, according to the agent's own choice, and the corresponding bridge rule is triggered.
\end{definition} 

Since agents' bridge rules are executed neither automatically nor simultaneously, we have to revise the definition of management function
with respect to the original definition of Section\rif{mMCS}.
First, notice that for each agent included in a DACmMCMAS the underlying logic $(KB_L; Cn_L; ACC_L)$ is such that:
$KB_L$ is composed of the global TBox plus the local ABoxes; $ACC_L$ is determined by the DRL-Lite semantics,
according to which elements of $Cn_L$ are computed. If an agent is equipped with
$n$ bridge rules, there will be $n$ operators in the agent's management base, one for each bridge rule, i.e., $OP = \{op_1,\ldots,op_n\}$.
Each of them which will at least make
the acquired knowledge compatible with the global TBox (possibly by means of global ontologies and/or
forms of meta-reasoning, cf., e.g., \cite{bcdl:jlc00,CostantiniMetaS02} for an overview). $F^{OP}_L$ is defined as in Section\rif{mMCS}, but instead of a single management
function there will now be $n$ management functions $\mathit{mng}_1,\ldots,\mathit{mng}_n$, again one per each bridge rule, each one with signature (for each $i$)
\[\mathit{mng}_i: {F^{OP}_L} \times KB^{L} \rightarrow  2^{KB_L} \setminus \emptyset\]
They can however be factorized within a single agent's management function with signature (as in MCSs)
\[\mathit{mng}_i: 2^{F^{OP}_L} \times KB^{L} \rightarrow  2^{KB_L} \setminus \emptyset\]
\noindent
which specializes into the $\mathit{mng}_i$s according to the bridge-rule head.

Whenever a bridge rule is triggered, its result is interpreted as an agent's generated event and is reacted to via a special ECA rule: this functioning is similar to \emph{internal events} in the DALI agent-oriented programming language \cite{jelia02,jelia04}.

\begin{definition}
\label{beca}
A \emph{bridge-update rule} has the form
\[\on A(\hat{x}) \tthen \beta(\hat{x})\]

where: $A(\hat{x})$ is the conclusion of exactly one bridge rule, and $\hat{x}$ a set of tuples representing the results of the application of the bridge rule; $\beta(\hat{x})$ specifies the operator, management function and actions to be applied to $\hat{x}$, which may imply querying the ABoxes of the agent and of the institutional agent, so as to re-elaborate the agent's ABox.
\end{definition} 

Actually, trigger and bridge-update rules can be seen as a special case of communicative and update rules of DACMASs, where the omitted recipient is implicitly assumed to be \emph{self}, i.e., the agent itself. The significant difference is that the involved event is not exactly an event as understood before, but is rather the result of a bridge rule.
Also, $\beta$ actions in bridge-update rules go beyond simple addition and deletion of facts performed by standard update rules.
We now need for DACmMCMAS agents an agent specification which is augmented w.r.t. that of DACMAS ones:

\begin{definition}
\label{agspec}
An agent specification is a tuple $\langle \mathit{sn},\Pi \rangle$, where $\mathit{sn}$ is
the agent specification name, and $\Pi$ is the set of rules characterizing the agent.
In particular,
$\Pi= \Pi_{\mathit{cu}} \cup \Pi_{\mathit{btu}} \cup \Pi_{\mathit{aux}}$, where $\Pi_{\mathit{cu}}$ is the set of communicative and update
rules, $\Pi_{\mathit{btu}}$ is the set of bridge, trigger and bridge-update
rules, and $\Pi_{\mathit{aux}}$ the set of the necessary auxiliary rules.
\end{definition} 

Notice that, though not explicitly mentioned, auxiliary rules where implicitly present also in the definition of DACMASs,
unless one considered all necessary auxiliary definitions as built-ins.

The definition of data state and of equilibria must be extended with respect to those provided in Section\rif{mMCS},
and not only because a data state now includes both contexts's and agents' sets of consequences. 
As mentioned, in MCSs a bridge rule is applied whenever it is applicable. This however does not in general imply that it is
applied only once, and that an equilibrium, once reached, lasts forever. In fact, contexts are in general able to
incorporate new data items from the external environment (which may include, as discussed in \cite{BrewkaEP14}, the input
provided by sensors). Therefore, a bridge rule is in principle re-evaluated whenever a new result can be obtained,
thus leading to evolving equilibria. In DACmMCMASs, there is the additional issue that for a bridge rule to be applied it
is not sufficient that it is applicable in the MCS sense, but it must also be triggered by a corresponding trigger rule.
Formally we have:

\begin{definition}
Let $M$ be a DACmMCMAS where $(C_1,\ldots,C_j)$ are the composing contexts and $(A_{j+1},\ldots,A_n)$ the composing agents, $j \geq 0, n>0$. A data state of $M$
is a tuple $S = (S_1,\ldots, S_n)$ such that each $S_i$ is
an element of $Cn_i$.
\end{definition}

\begin{definition}
Lat $S = (S_1,\ldots, S_n)$ be a data set for a DACmMCMAS $M$ where $(C_1,\ldots,C_j)$ are the composing contexts and $(A_{j+1},\ldots,A_n)$ are the composing agents.
A bridge rule $\rho$ occurring in each composing context or agent is \emph{potentially applicable} in $S$ iff $S$ entails its body. For contexts, entailment is the same as in MCSs. For agents, entailment implies that all queries in the rule body succeed w.r.t. $S$. A bridge rule is \emph{applicable in a 
context} whenever it is potentially applicable.  A bridge rule with head $A(\hat{y})$ is \emph{applicable in an
agent} $A_j$ whenever it is potentially applicable
and there exists a trigger rule of the form $Q(\hat{x}) \enables A(\hat{y})$ in the specification of $A_j$ such that $Cn_j \models Q(\hat{x})$. Let $\mathit{app}(S)$ be the set of bridge rules which are applicable in data state $S$.
\end{definition}

Desirable data states are those where each $S_i$ is acceptable according to $ACC_i$.

\begin{definition}
A data state is an
equilibrium iff, for $1 \leq i \leq n$, there exists $kb_i' =  mng_i(\mathit{app}(S),kbi)$ such that
$S_i \in ACC_i(kb_i')$.
\end{definition}

We will now re-elaborate an example taken from \cite{MontaliCG14}, where a potential buyer queries the institutional agent
in order to obtain the name of the seller (of some goods) and then sends to the seller a registration request,
supposing that registration is required in order to become a customer. The communication rule from \cite{MontaliCG14} is in particular
the following, where \emph{sel} is the variable which is instantiated by the query to the seller's name. The query
 in fact asks $\mathit{inst}$ for the specification name $\mathit{Spec}$ of $\mathit{seller}$.
 
\[\mathit{Spec@inst}(\mbox{\emph{sel}},\mathit{seller})\ \enables \mathit{REQ\_REG}\ \tto\ \mbox{\emph{sel}}\]

We assume instead that before sending the registration request, the potential buyer intends to
verify the reliability of the seller. To do so, the above rule is substituted by the following trigger rule,
which looks almost identical unless for the fact that, instead of sending an event, a bridge rule is invoked
by mentioning is head.

\[\mathit{Spec@inst}(\mbox{\emph{sel}}),\mathit{seller})\ \enables \mathit{verify(\mbox{\emph{sel}})}\]

\noindent
The bridge rule might involve consulting a list of trusted companies and reviewing other user's opinion.

\[\begin{array}{l}
\mathit{verify(\mbox{\emph{sel}})} \determinedby\\ 
\tbl \mathit{trusted(\mbox{\emph{sel}})\,:\,trusted\_companies\_directory},\\
\tbl \no \mathit{bad\_rating(\mbox{\emph{sel}})\,:\,user\_forum} 
\end{array}
\]

If the verification is successful, which implies that $\mathit{verify(\mbox{\emph{sel}})}$ becomes true
as the body is implied by the current data state, then the seller can be added to the agent's ABox as reliable,
and the registration request can actually be sent. This is achieved via the following bridge-update rule

\[\on \mathit{verify(\mbox{\emph{sel}})}\ \tthen\ \mathit{add(verified(\mbox{\emph{sel}}))}\]

\noindent
combined with a corresponding communication rule, whose head corresponds to a query of the agent
to its own ABox. 

\[\mathit{verified(\mbox{\emph{sel})}} \enables \mathit{REQ\_REG}\ \tto\ \mbox{\emph{sel}}\]

Bridge-update rules and communication rules are detached, as the former can subsist independently
of the latter. If communication occured in the bridge-update rule, this would establish an
unnecessary constraint about communication always occurring, and occurring immediately after 
information acquisition.

\subsection{Properties of DACmMCMASs}
\label{props}

In the terminology of \cite{BrewkaEF11}, we require all management functions (both those related to agents and those related to contexts)
to be \emph{local consistency (lc-)preserving}. 
We thus obtain the following, as a consequence of Proposition 2 in \cite{BrewkaEF11}:

\begin{proposition}
Let D be a DACmMCMAS such that all management functions associated to 
the composing agents and contexts are lc-preserving.
Then $D$ is locally consistent.
\end{proposition}

The execution semantics of a DACmMCMAS can be defined by extending the \emph{transition system} defined for DACMAS in \cite{MontaliCG14}, Section 4. We omit the full extended definition for lack of space. It should suffice however to say that the transition system construction (Figure 1 in \cite{MontaliCG14}) must be modified: (i) in steps 2-24, which manage rules, so as to encompass bridge rules; (ii) in steps 30-37, which manage actions, so as to encompass $\beta$ actions in bridge-update rules. 

This allows us to extend to DACmMCMASs the nice results provided in \cite{MontaliCG14} for DACMASs about verification using $\mu$-calculus (which is a powerful temporal logic used for model checking of finite-state transition systems, able to express both linear-time temporal logics such as LTL and branching-time temporal logics such as CTL and its variants). In particular, in \cite{MontaliCG14} a variant of $\mu$-calculus is adopted, that we denote as $\mu_{ECQ_{l}}$ (for the formal definition of this variant, the reader may refer to \cite{CalvaneseGMP13} and to the references therein). In Section 5.1 of \cite{MontaliCG14}, decidability of verification of DACMASs is proved by assuming that they are \emph{state bounded}, in the sense that for each agent in a DACMAS ther exists a bound on the number of data items simultaneously stored in its ABOX. 
In our context, we can assume state-boundedness for both agents' ABoxes and contexts' data instances:
i.e, we assume that also the number of data items simultaneously stored
in a context stays within the same bound as agents' ABoxes. Thus, verifiability properties
of DACMASs (Theorem 5.1 in \cite{MontaliCG14}) still holds for DACmMCMASs.
The proof of the theorem remains substantially the same: 
in \cite{MontaliCG14} it is assumed to unify all the ABoxes of the composing agents into a single relational database where tuples have an additional argument denoting the name of the agent the tuple comes from. We can assume to include in this relational database also data belonging to contexts, with an additional argument denoting the context name. 

\section{Concluding Remarks}
\label{conclusions}

In this paper we have extended DACMAS, which is a formalization of ontology-based and commitment-based multi-agent systems. Our extension allows a system to include not only agents but also external contexts. The objective is that of modeling real-world situations where agents not only interact among themselves, but also consult external heterogeneous data- and knowledge-based to extract useful information. In this way, both single agents and the overall system is able to evolve by incorporating new data/knowledge. To do so, we have merged DACMASs with managed multi-context systems (mMCS), which model data and knowledge exchange among heterogeneous sources in a controlled way. We thus obtain a more general formalization, called DACmMCMAS, which retains the desirable properties of both approaches.
Similarly to what the authors argue for DACMAS, instances of DACmMCMAS are readily implementable via standard publicly available technologies. 

We may notice however that a-priori formal verification may not always be practically possible, even for DACMASs. In fact, decidability of verification relies upon the assumption that agents's ABoxes are publicly available and can be in principle merged together. This assumption may certainly apply to many practical cases. In general however, agents are supposed to be wishing to keep some of their information private and unknown to other agents and to a potential third-party. Even more so in the case of DACmMCMAS: sometimes contexts will be part of the same overall organization, which would make contexts' contents available thus enabling verification. Sometimes however they can be fully external to the MAS. For these reason, we believe that a-priori verification techniques might be profitably complemented by run-time verification techniques, such as, e.g., those discussed in \cite{Costantini12,CostantiniG14}.
}\end{sloppypar}


\begin{thebibliography}{10}

\bibitem{ArtikisP08}
Artikis, A., Pitt, J.V.:
\newblock Specifying open agent systems: A survey.
\newblock In Artikis, A., Picard, G., Vercouter, L., eds.: Engineering
  Societies in the Agents World IX, 9th International Workshop, ESAW 2008,
  Revised Selected Papers. (2008)  29--45

\bibitem{ACL-Manifesto2013}
Chopra, A.K., Artikis, A., Bentahar, J., Colombetti, M., Dignum, F., Fornara,
  N., Jones, A.J.I., Singh, M.P., Yolum, P.:
\newblock Research directions in agent communication.
\newblock ACM TIST \textbf{4}(2) (2013) ~20

\bibitem{Singh91}
Singh, M.P.:
\newblock Towards a formal theory of communication for multi-agent systems.
\newblock In Mylopoulos, J., Reiter, R., eds.: Proceedings of the 12th
  International Joint Conference on Artificial Intelligence, Morgan Kaufmann
  (1991)  69--74

\bibitem{Singh07}
Singh, M.P.:
\newblock Formalizing communication protocols for multiagent systems.
\newblock In Veloso, M.M., ed.: IJCAI 2007, Proceedings of the 20th
  International Joint Conference on Artificial Intelligence. (2007)  1519--1524

\bibitem{Singh00}
Singh, M.P.:
\newblock A social semantics for agent communication languages.
\newblock In Dignum, F., Greaves, M., eds.: Issues in Agent Communication.
  Volume 1916 of Lecture Notes in Computer Science., Springer (2000)  31--45

\bibitem{Singh12}
Singh, M.P.:
\newblock Commitments in multiagent systems: Some history, some confusions,
  some controversies, some prospects.
\newblock In Paglieri, F., Tummolini, L., Falcone, R., Miceli, M., eds.: The
  Goals of Cognition. Essays in Honor of Cristiano Castelfranchi, College
  Publications, London (2012)  601–626

\bibitem{YolumS01}
Yolum, P., Singh, M.P.:
\newblock Commitment machines.
\newblock In Meyer, J.J.C., Tambe, M., eds.: Intelligent Agents VIII, 8th
  International Workshop, ATAL 2001, Revised Papers. Volume 2333 of Lecture
  Notes in Computer Science., Springer (2001)  235--247

\bibitem{Singh99}
Singh, M.P.:
\newblock An ontology for commitments in multiagent systems.
\newblock Artif. Intell. Law \textbf{7}(1) (1999)  97--113

\bibitem{REC2011}
Torroni, P., Chesani, F., Mello, P., Montali, M.:
\newblock A retrospective on the reactive event calculus and commitment
  modeling language.
\newblock In Sakama, C., Sardi{\~n}a, S., Vasconcelos, W., Winikoff, M., eds.:
  Declarative Agent Languages and Technologies IX - 9th International Workshop,
  DALT 2011, Revised Selected and Invited Papers. Volume 7169 of Lecture Notes
  in Computer Science., Springer (2012)  120--127

\bibitem{REC2012}
Bragaglia, S., Chesani, F., Mello, P., Montali, M., Torroni, P.:
\newblock Reactive event calculus for monitoring global computing applications.
\newblock In Artikis, A., Craven, R., Cicekli, N.K., Sadighi, B., Stathis, K.,
  eds.: Logic Programs, Norms and Action - Essays in Honor of Marek J. Sergot
  on the Occasion of His 60th Birthday. Volume 7360 of Lecture Notes in
  Computer Science., Springer (2012)  123--146

\bibitem{ChesaniMMT13}
Chesani, F., Mello, P., Montali, M., Torroni, P.:
\newblock Representing and monitoring social commitments using the event
  calculus.
\newblock Autonomous Agents and Multi-Agent Systems \textbf{27}(1) (2013)
  85--130

\bibitem{BaldoniBCDPS09}
Baldoni, M., Baroglio, C., Chopra, A.K., Desai, N., Patti, V., Singh, M.P.:
\newblock Choice, interoperability, and conformance in interaction protocols
  and service choreographies.
\newblock In Sierra, C., Castelfranchi, C., Decker, K.S., Sichman, J.S., eds.:
  8th International Joint Conference on Autonomous Agents and Multiagent
  Systems AAMAS 2009, IFAAMAS (2009)  843--850

\bibitem{MarengoBBCPS11}
Marengo, E., Baldoni, M., Baroglio, C., Chopra, A.K., Patti, V., Singh, M.P.:
\newblock Commitments with regulations: reasoning about safety and control in
  regula.
\newblock In Sonenberg, L., Stone, P., Tumer, K., Yolum, P., eds.: 10th
  International Conference on Autonomous Agents and Multiagent Systems AAMAS
  2011, IFAAMAS (2011)  467--474

\bibitem{BaldoniBMP13}
Baldoni, M., Baroglio, C., Marengo, E., Patti, V.:
\newblock Constitutive and regulative specifications of commitment protocols: A
  decoupled approach.
\newblock ACM TIST \textbf{4}(2) (2013) ~22

\bibitem{Singh13}
Gerard, S.N., Singh, M.P.:
\newblock Evolving protocols and agents in multiagent systems.
\newblock In Gini, M.L., Shehory, O., Ito, T., Jonker, C.M., eds.:
  International conference on Autonomous Agents and Multi-Agent Systems, AAMAS
  '13. (2013)  997--1004

\bibitem{FIPA2003}
{Foundation for Intelligent Physical Agents}:
\newblock {FIPA Interaction Protocolo Specifications} (2003)

\bibitem{rao}
Rao, A.S., Georgeff, M.P.:
\newblock Modeling agents within a {B}{D}{I}-architecture.
\newblock In Fikes, R., Sandewall, E., eds.: Proceedings of International
  Conference on Principles of Knowledge Representation and Reasoning (KR),
  Cambridge, Massachusetts, Morgan Kaufmann (1991)

\bibitem{RG91}
Rao, A.S., Georgeff, M.:
\newblock Modeling rational agents within a bdi-architecture.
\newblock In: Proceedings of the Second Int. Conf. on Principles of Knowledge
  Representation and Reasoning (KR'91), Morgan Kaufmann (1991)  473--484

\bibitem{ForCol2009}
Fornara, M., Colombetti, N.:
\newblock Specifying artificial institutions in the event calculus.
\newblock In Dignum, V., ed.: Handbook of Research on Multi-Agent Systems:
  Semantics and Dynamics of Organizational Models.
\newblock IGI Global (2009)  335--366 Chapter 14.

\bibitem{SinghMASC}
Chopra, A.K., Singh, M.P.:
\newblock Agent communication.
\newblock In Weiss, G., ed.: Multiagent Systems, 2nd edition.
\newblock MIT Press (2013)

\bibitem{MontaliCG14}
Montali, M., Calvanese, D., {De Giacomo}, G.:
\newblock Specification and verification of commitment-regulated data-aware
  multiagent systems.
\newblock In: Proceedings of AAMAS 2014, also in Proceedings of the 29th
  Italian Conference on Computational Logic,
  http://ceur-ws.org/Vol-1195/long6.pdf. (2014)

\bibitem{Stirling01}
Stirling, C.:
\newblock Modal and Temporal Properties of Processes.
\newblock Texts in Computer Science. Springer (2001)

\bibitem{CalvaneseGMP13}
Calvanese, D., {De Giacomo}, G., Montali, M., Patrizi, F.:
\newblock Verification and synthesis in description logic based dynamic systems
  (abridged version).
\newblock In Faber, W., Lembo, D., eds.: Web Reasoning and Rule Systems - 7th
  International Conference, RR 2013. Volume 7994 of Lecture Notes in Computer
  Science., Springer (2013)

\bibitem{DL2003handbook}
Baader, F., Calvanese, D., McGuinness, D.L., Nardi, D., Patel{-}Schneider,
  P.F.:
\newblock The description logic handbook: Theory, implementation, and
  applications.
\newblock Cambridge University Press (2003)

\bibitem{BrewkaE07}
Brewka, G., Eiter, T.:
\newblock Equilibria in heterogeneous nonmonotonic multi-context systems.
\newblock In: Proceedings of the 22nd AAAI Conference on Artificial
  Intelligence, AAAI Press (2007)  385--390

\bibitem{BrewkaEF11}
Brewka, G., Eiter, T., Fink, M.:
\newblock Nonmonotonic multi-context systems: A flexible approach for
  integrating heterogeneous knowledge sources.
\newblock In Balduccini, M., Son, T.C., eds.: Logic Programming, Knowledge
  Representation, and Nonmonotonic Reasoning - Essays Dedicated to Michael
  Gelfond on the Occasion of His 65th Birthday. Volume 6565 of Lecture Notes in
  Computer Science., Springer (2011)  233--258

\bibitem{BrewkaEFW11}
Brewka, G., Eiter, T., Fink, M., Weinzierl, A.:
\newblock Managed multi-context systems.
\newblock In Walsh, T., ed.: IJCAI 2011, Proceedings of the 22nd International
  Joint Conference on Artificial Intelligence, IJCAI/AAAI (2011)  786--791

\bibitem{lpneg:survey}
Apt, K.R., Bol, R.:
\newblock Logic programming and negation: A survey.
\newblock The Journal of Logic Programming \textbf{19-20} (1994)  9--71

\bibitem{bcdl:jlc00}
Barklund, J., Dell'Acqua, P., Costantini, S., Lanzarone, G.A.:
\newblock Reflection principles in computational logic.
\newblock J. of Logic and Computation \textbf{10}(6) (2000)  743--786

\bibitem{CostantiniMetaS02}
Costantini, S.:
\newblock Meta-reasoning: A survey.
\newblock In: Computational Logic: Logic Programming and Beyond, Essays in
  Honour of Robert A. Kowalski, Part II. Volume 2408 of Lecture Notes in
  Computer Science.
\newblock Springer (2002)

\bibitem{jelia02}
Costantini, S., Tocchio, A.:
\newblock A logic programming language for multi-agent systems.
\newblock In: Logics in Artificial Intelligence, Proceedings of the 8th Europ.
  Conf.,JELIA 2002. LNAI 2424, Springer-Verlag, Berlin (2002)

\bibitem{jelia04}
Costantini, S., Tocchio, A.:
\newblock The {{DALI}} logic programming agent-oriented language.
\newblock In: Logics in Artificial Intelligence, Proceedings of the 9th
  European Conference, Jelia 2004. LNAI 3229, Springer-Verlag, Berlin (2004)

\bibitem{BrewkaEP14}
Brewka, G., Ellmauthaler, S., P{\"u}hrer, J.:
\newblock Multi-context systems for reactive reasoning in dynamic environments.
\newblock In Schaub, T., ed.: ECAI 2014, Proceedings of the 21st European
  Conference on Artificial Intelligence, IJCAI/AAAI (2014)

\bibitem{Costantini12}
Costantini, S.:
\newblock Self-checking logical agents.
\newblock In Osorio, M., Zepeda, C., Olmos, I., Carballido, J.L., Ram\'{\i}rez,
  R.C.M., eds.: Proceedings of the Eighth Latin American Workshop on Logic,
  Languages, Algorithms and New Methods of Reasoning LA-NMR 2012. Volume 911 of
  CEUR Workshop Proceedings., CEUR-WS.org (2012)  3--30 Invited Paper, Extended
  Abstract in Proc. of AAMAS 2013, Twelfth Intern. Conf. on Autonomous Agents
  and Multi-Agent Systems.

\bibitem{CostantiniG14}
Costantini, S., Gasperis, G.D.:
\newblock Runtime self-checking via temporal (meta-)axioms for assurance of
  logical agent systems.
\newblock In Bulling, N., van~der Hoek, W., eds.: Proceedings of LAMAS 2014,
  7th Workshop on Logical Aspects of Multi-Agent Systems, held at AAMAS 2014,
  14th Intern. Conf. on Autonomous Agents and Multi-Agent Systems. (2014)
  241--255 also in Proc, of the 29th Italian Conference on Computational Logic.

\end{thebibliography}
\end{document}